\documentclass[11pt,a4paper]{article}
\usepackage[hyperref]{naaclhlt2018}
\usepackage{times}
\usepackage{latexsym}
\usepackage{booktabs}
\usepackage{amsmath}
\usepackage{amssymb}
\usepackage{booktabs}
\usepackage[]{graphicx}
\graphicspath{{figures/}}
\usepackage{tikz}
\usepackage{tikz-dependency}
\usetikzlibrary{arrows}
\usepackage{xspace}

\usepackage{url}

\aclfinalcopy 

\title{Exploiting Semantics in Neural Machine Translation with \\ Graph Convolutional Networks}

\author{
 Diego Marcheggiani$^{1,2}$ \hspace{1cm} Jasmijn Bastings$^{2}$\hspace{1cm} Ivan Titov$^{1,2}$    \\
 $^1$ILCC, School of Informatics, University of Edinburgh \\
 $^2$ILLC,  University of Amsterdam  \\
    {\tt marcheggiani@uva.nl} \hspace{0.3cm} {\tt bastings@uva.nl} \hspace{0.3cm} {\tt ititov@inf.ed.ac.uk}
 }

\date{}

\begin{document}
\maketitle
\begin{abstract}
  Semantic representations have long been argued as potentially useful for enforcing meaning preservation and improving generalization performance of machine translation methods. 
  In this work, we are the first to incorporate information about predicate-argument structure of source sentences (namely, semantic-role representations) into neural machine translation. We use Graph Convolutional Networks (GCNs) to inject a semantic bias into sentence encoders and achieve improvements in BLEU scores over the linguistic-agnostic and syntax-aware versions on the English--German language pair.  

\end{abstract}

\section{Introduction}

It has long been argued that semantic representations may provide a useful linguistic bias to machine translation systems~\cite{weaver1955translation,bar1960present}. Semantic
representations provide an abstraction which can generalize over different surface realizations of the same underlying `meaning'. 
Providing this information to a machine translation system, can, in principle, improve meaning preservation and boost generalization performance.

Though incorporation of semantic information into traditional statistical machine translation has been an active research topic~(e.g., \cite{baker-etal:2012:CL,DBLP:conf/coling/LiuG10,DBLP:conf/naacl/WuF09,DBLP:conf/acl/BazrafshanG13,DBLP:conf/wmt/AzizRS11,C12-1083}), we are not aware of any previous work considering semantic structures in neural machine translation (NMT). In this work, we aim to fill this gap by showing how information about predicate-argument structure of source sentences can be integrated into standard attention-based NMT models~\cite{bahdanau15iclr}.

\begin{figure}
\begin{center}
\includegraphics[width=1 \columnwidth, height=3.65cm, keepaspectratio]{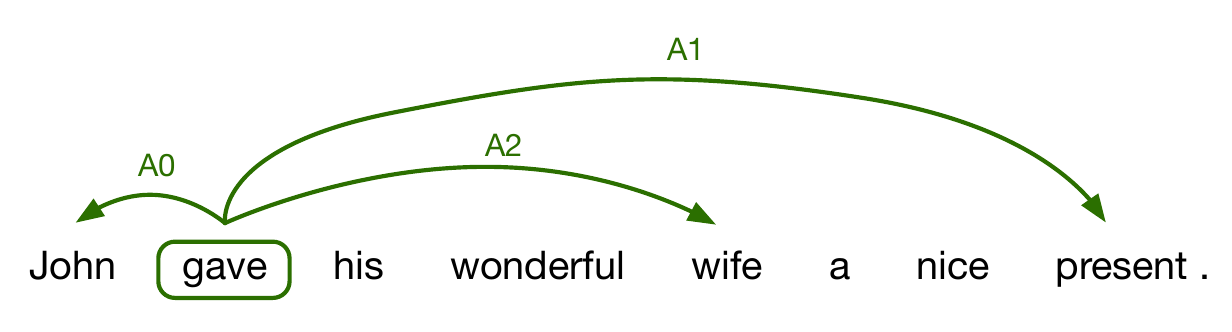}
\vspace{-3ex}
\caption{ 
An example sentence annotated with a semantic-role representation.
\label{fig:srl-example}} 
\vspace{-2.5ex}
\end{center}
\end{figure}

We consider PropBank-style \cite{DBLP:journals/coling/PalmerKG05} semantic role structures, or more specifically their dependency versions~\cite{DBLP:conf/conll/SurdeanuJMMN08}. 
The semantic-role representations mark semantic arguments of predicates in a sentence and categorize them according  to their {\it semantic roles}. 
Consider Figure \ref{fig:srl-example}, the predicate {\it gave} has three arguments:\footnote{We slightly abuse the terminology: formally these are syntactic heads of arguments rather than arguments.} \textit{John} (semantic role {\tt A0}, `the giver'), \textit{wife} (\texttt{A2}, `an entity given to') and \textit{present} (\texttt{A1}, `the thing given'). Semantic roles capture commonalities between different realizations of the same underlying predicate-argument structures. 
For example, {\it present} will still be {\tt A1} in sentence {\it``John gave a nice present to his wonderful wife''}, despite different surface 
forms of the two sentences. We hypothesize that semantic roles can be especially beneficial in NMT, as `argument switching' (flipping arguments corresponding to different roles) is one of frequent and severe mistakes made by NMT systems \cite{DBLP:conf/emnlp/IsabelleCF17}.

There is a limited amount of work on incorporating graph structures into neural sequence models. 
Though, unlike semantics in NMT, syntactically-aware NMT has been a relatively hot topic recently, with a number of approaches claiming improvements from using treebank syntax ~\cite{sennrich2016linguistic,eriguchi2016treetoseq,DBLP:conf/wmt/NadejdeRSDJKB17,bastings-EtAl:2017:EMNLP2017,aharonigoldberg2017stringtotree}, our graphs are different from syntactic structures. Unlike syntactic dependency graphs, they are not trees and thus cannot be processed in a bottom-up fashion as in \newcite{eriguchi2016treetoseq} or easily linearized as in \newcite{aharonigoldberg2017stringtotree}. Luckily, the modeling approach of \newcite{bastings-EtAl:2017:EMNLP2017} does not make any assumptions about the graph structure, and thus we build on their method.

\newcite{bastings-EtAl:2017:EMNLP2017} used Graph Convolutional Networks (GCNs) to encode syntactic structure. GCNs were originally proposed by 
 \newcite{kipf2016semigraphconv} and  modified to handle labeled and automatically predicted (hence noisy) syntactic dependency graphs by ~\newcite{marcheggiani-titov:2017:srlgcn}. Representations of nodes (i.e. words in a sentence) in GCNs are directly influenced by representations of their neighbors in the graph. The form of influence (e.g., transition matrices and parameters of gates) are learned in such a way as to benefit the end task (i.e. translation). These linguistically-aware word representations are used within a neural encoder.  Although recent research has shown that neural architectures are able to learn some linguistic phenomena without explicit linguistic supervision \cite{linzen-dupoux-goldberg:2016:tacllstmsyntax,NIPS2017_7181}, informing word representations with linguistic structures can provide a useful inductive bias.

We apply GCNs to the semantic dependency graphs and experiment on the  English--German language pair (WMT16). 
We observe an improvement over the semantics-agnostic baseline (a BiRNN encoder; 23.3 vs 24.5 BLEU). 
As we use exactly the same modeling approach as in the syntactic method of \newcite{bastings-EtAl:2017:EMNLP2017}, we can easily compare the influence of the types of linguistic structures (i.e., syntax vs. semantics). 
We observe that when using full WMT data we obtain better results with semantics than with syntax (23.9 BLEU for syntactic GCN).
Using syntactic and semantic GCN together, we obtain a further gain (24.9 BLEU) that suggests the complementarity of syntax and semantics.

\section{Model}

\subsection{Encoder-decoder Models}

We use a standard attention-based encoder-decoder model \cite{bahdanau15iclr} as a starting point for constructing our model.
In encoder-decoder models, the encoder takes as input the source sentence $\mathbf{x}$ and calculates a representation of each word $x_t$ in $\mathbf{x}$. 
The decoder outputs a translation $\mathbf{y}$  relying on the representations of the source sentence.
Traditionally, the encoder is parametrized as a Recurrent Neural Network (RNN), but other architectures have also been successful, such as Convolutional Neural Networks (CNN) \cite{gehring2016convolutional} and hierarchical self-attention models \cite{NIPS2017_7181}, among others.
In this paper we experiment with RNN and CNN encoders. 
We explore the benefits of incorporating information about semantic-role structures into such encoders.

More formally, RNNs \cite{elman1990finding} can be defined as a function $ \mathrm{RNN}(x_{1:t})$ that calculates the hidden representation $h_t$ of a word $x_t$ based on its left context.
Bidirectional RNNs use two RNNs: one runs in the forward direction and another one in the backward direction.
The forward $ \mathrm{RNN}(x_{1:t})$ represents the left context of word $x_t$, whereas the backward $ \mathrm{RNN}(x_{n:t})$ computes a representation of the right context. The two representations are concatenated in order to incorporate information about the entire sentence:  $$ \mathrm{BiRNN}(\mathbf{x},t) =  \mathrm{RNN}(x_{1:t})  \circ  \mathrm{RNN}(x_{n:t}).$$
In contrast to BiRNNs, CNNs \cite{lecun-01a} calculate a representation of a word $x_t$ by considering a window of words $w$ around $x_t$, such as $$ \mathrm{CNN}(\mathbf{x}, t, w) = f(x_{t-\lfloor w/2 \rfloor}, .., x_t, .., x_{t+\lfloor w/2 \rfloor}),$$
where $f$ is usually an affine transformation followed by a nonlinear function.

Once the sentence has been encoded, the decoder takes as input the induced sentence representation and generates the target sentence $\mathbf{y}$.
The target sentence $\mathbf{y}$ is predicted word by word using an RNN decoder. 
At each step, the decoder calculates the probability of generating a word $y_t$ conditioning on a context vector $\mathbf{c_t}$ and the previous state of the RNN decoder. 
The context vector $\mathbf{c_t}$ is calculated based on the representation of the source sentence computed by the encoder, using an attention mechanism \cite{bahdanau15iclr}. 
Such a model is trained end-to-end on a parallel corpus to maximize the conditional likelihood of the target sentences.

\subsection{Graph Convolutional Networks}
\begin{figure}
\begin{center}
\includegraphics[width=7.8cm, height=8cm, keepaspectratio]{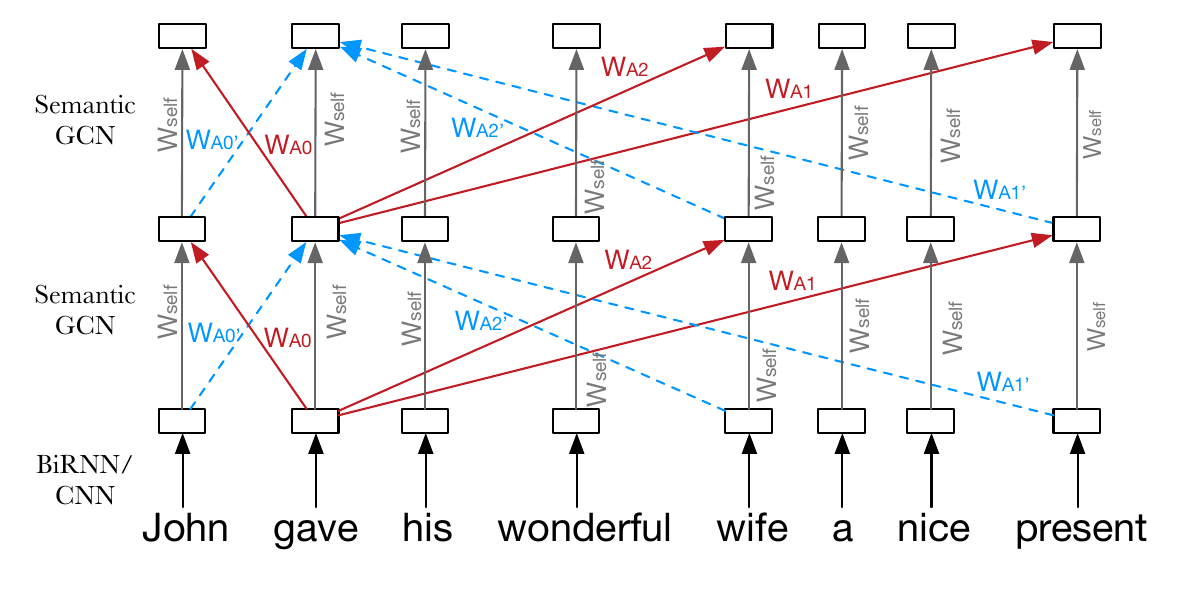}
\caption{ 
\label{fig:sem-gcn} Two layers of semantic GCN on top of a (not shown) BiRNN or CNN encoder.
} 
\end{center}
\end{figure}

\begin{table}[h]
\centering
\begin{tabular}{@{}lrr@{}} \toprule
 & BiRNN & CNN\\ 
\midrule
Baseline \small{\cite{bastings-EtAl:2017:EMNLP2017}}	& 14.9  &  12.6\\
\,\,+Sem	&  15.6 & 13.4\\
\,\,+Syn \small{\cite{bastings-EtAl:2017:EMNLP2017}} & 16.1 & 13.7 \\
\,\,+Syn + Sem	 & 15.8 & 14.3 \\
\bottomrule
\end{tabular}
\caption{Test BLEU, En--De, News Commentary.}
\label{tab:test-small}
\end{table}

\begin{table}[h]
\centering
\begin{tabular}{@{}lr@{}} \toprule
 & BiRNN  \\
\midrule
Baseline \small{\cite{bastings-EtAl:2017:EMNLP2017}}	& 23.3 \\ 
\,\,+Sem	&  24.5  \\  
\,\,+Syn \small{\cite{bastings-EtAl:2017:EMNLP2017}} & 23.9 \\ 
\,\,+Syn + Sem & 24.9 \\ 
\bottomrule
\end{tabular}
\caption{Test BLEU, En--De, full WMT16.}
\label{tab:test-full}
\end{table}

Graph neural networks are a family of neural architectures \cite{scarselli2009graph,DBLP:conf/icml/GilmerSRVD17} specifically devised to induce representation of nodes in a graph relying on its graph structure.
Graph convolutional networks (GCNs) belong to this family.
While GCNs were introduced for modeling undirected unlabeled graphs \cite{kipf2016semigraphconv}, in this paper we use a formulation of GCNs for labeled directed graphs, where the direction and the label of an edge are incorporated.
In particular, we follow the formulation of \newcite{marcheggiani-titov:2017:srlgcn} and \newcite{bastings-EtAl:2017:EMNLP2017} for syntactic graphs and apply it to dependency-based semantic-role structures \cite{DBLP:conf/conll/HajicCJKMMMNPSSSXZ09} (as in  Figure \ref{fig:srl-example}).

More formally, consider a directed graph $\mathcal{G} = (\mathcal{V}, \mathcal{E})$, where $\mathcal{V}$ is a set of nodes, and $\mathcal{E}$ is a set of edges. 
Each node $v \in \mathcal{V}$ is represented by a feature vector $\mathbf{x}_v \in \mathbb{R}^d$, where $d$ is the latent space dimensionality.
The GCN induces a new representation $\mathbf{h}'_v \in \mathbb{R}^d$ of a node $v$ while relying on representations $\mathbf{h}_u$ of its neighbors:
\begin{align*}
\mathbf{h}'_v\!\!=&\rho\Big(\!\!\!\sum_{u \in \mathcal{N}(v)} \!\!\! g_{u, v} \big(W_{dir(u, v)} \, \mathbf{h}_u + \mathbf{b}_{lab(u, v)}\big)\Big),
\end{align*}
where $\mathcal{N}(v)$ is the set of neighbors of $v$, $W_{dir(u, v)} \in \mathbb{R}^{d \times d}$ is a direction-specific parameter matrix. 
There are three possible directions ($dir(u, v) \in \{in, out, loop\}$): self-loop edges were added in order to ensure that the initial representation of node $\mathbf{h}_v$ directly affects its new representation $\mathbf{h}'_v$. 
The vector $\mathbf{b}_{lab(u, v)} \in \mathbb{R}^d$ is an embedding of a semantic role label of the edge $(u,v)$ (e.g., {\tt A0}). 
The functions $g_{u, v}$ are scalar gates which weight the importance of each edge.
Gates are particularly useful when the graph is predicted and thus may contain errors, i.e., wrong edges.
In this scenario gates can down weight the influence of such edges. 
$\rho$ is a non-linearity (ReLU).\footnote{Refer to \newcite{marcheggiani-titov:2017:srlgcn} and \newcite{bastings-EtAl:2017:EMNLP2017} for further details.}

As with CNNs, GCN layers can be stacked in order to incorporate higher order neighborhoods. 
In our experiments, we used GCNs on top of a standard BiRNN encoder and a CNN encoder (Figure \ref{fig:sem-gcn}). 
In other words, the initial representations of words fed into GCN were either RNN states or CNN representations.

\section{Experiments} 
We experimented with the English-to-German WMT16 dataset ($\sim$4.5 million sentence pairs for training).
We use its subset, News Commentary v11, for development and additional experiments ($\sim$226.000 sentence pairs).
For all these experiments, we use \texttt{newstest2015} and \texttt{newstest2016} as a validation and test set, respectively.

We parsed the English partitions of these datasets with a syntactic dependency parser \cite{P16-1231} and dependency-based semantic role labeler \cite{marcheggiani-frolov-titov:2017:srl}.
We constructed the English vocabulary by taking all words with frequency higher than three, while for German we used byte-pair encodings (BPE) \cite{sennrich2016subword}.
All hyperparameter selection was performed on the validation set (see Appendix \ref{sec:hyperparameters}).
We measured the performance of the models with (cased) BLEU scores \cite{papineni2002bleu}.
The settings and the framework (Neural Monkey \cite{NeuralMonkey:2017}) used for experiments are the ones used in \newcite{bastings-EtAl:2017:EMNLP2017}, which we use as baselines. As RNNs, we use GRUs~\cite{cho14emnlp}.

We now discuss the impact that different architectures and linguistic information have on the translation quality. 

\subsection{Results and Discussion}

First, we start with experiments with the smaller News Commentary training set (See Table \ref{tab:test-small}).  
As in \newcite{bastings-EtAl:2017:EMNLP2017}, we used the standard attention-based encoder-decoder model as a baseline.

We tested the impact of semantic GCNs when used on top of CNN and BiRNN encoders. As expected, BiRNN results are stronger than CNN ones.
In general, for both encoders we observe the same trend: using semantic GCNs leads to an improvement over the baseline model. 
The improvements is 0.7 BLEU for BiRNN and 0.8 for CNN. 
This is slightly surprising as the potentially non-local semantic information should in principle be more beneficial within a less powerful and local CNN encoder. 
The syntactic GCNs  \cite{bastings-EtAl:2017:EMNLP2017} appear stronger than semantic GCNs. 
As exactly the same model and optimization are used for both GCNs, the differences should be due to the type of linguistic representations used.\footnote{Note that the SRL system we use~\cite{marcheggiani-frolov-titov:2017:srl} does not use syntax and is faster than the syntactic parser of \newcite{P16-1231}, so semantic GCNs may still be preferable from the engineering perspective even in this setting.}
When syntactic and semantic GCNs are used together, we observe a further improvement with respect to the semantic GCN model, and a substantial improvement with respect to the syntactic GCN model with a CNN encoder.

Now we turn to the full WMT experiments. 
Though we expected that the linguistic bias should more valuable in a resource-poor setting, the improvement from using semantic-role structures is larger here (+1.2 BLEU). 
It is surprising but perhaps more data is beneficial for accurately modeling influence of semantics on the translation task. 
Interestingly, the semantic GCN now outperforms the syntactic one by 0.6 BLEU. Again, it is hard to pinpoint exact reasons for this. One may speculate though that, given enough data, RNNs were able to capture syntactic dependency and thus reducing the benefits from using treebank syntax,  whereas (often less local and harder) semantic dependencies were more complementary. 
Finally, when syntactic and semantic GCN are trained together, we obtain a further improvement reaching 24.9 BLEU.
These results suggest that syntactic and semantic dependency structures are complementary information when it comes to translation.

\begin{table}[h]
\centering
\begin{tabular}{@{}lrr@{}} 
\toprule
& BiRNN & CNN \\
\midrule
Baseline \small{\cite{bastings-EtAl:2017:EMNLP2017}}	& 14.1 & 12.1\\
\midrule
\,\,+Sem (1L)	& 14.3 & 12.5\\
\,\,+Sem (2L)	& 14.4 &12.6\\
\,\,+Sem (3L)	& 14.4	&12.7\\
\,\,+Syn (2L) \small{\cite{bastings-EtAl:2017:EMNLP2017}} & 14.8 & 13.1\\
\midrule
\,\,+SelfLoop (1L)			& 14.1	& 12.1 \\
\,\,+SelfLoop (2L)			& 14.2	& 11.5\\
\toprule
\,\,+SemSyn (1L)			& 14.1	& 12.7	\\
\,\,+Syn (1L) + Sem (1L)	& 14.7	& 12.7\\
\,\,+Syn (1L) + Sem (2L)	& 14.6	& 12.8 \\
\,\,+Syn (2L) + Sem (1L)	& 14.9	& 13.0	 \\
\,\,+Syn (2L) + Sem (2L)	& 14.9 & 13.5\\
\bottomrule
\end{tabular}
\caption{Validation BLEU, News commentary only}
\label{tab:dev-ende}
\end{table}

\begin{table*}[h]
\centering
\begin{tabular}{@{}ll@{}} 
\toprule
 & \includegraphics[width=1 \columnwidth, height=3.65cm, keepaspectratio]{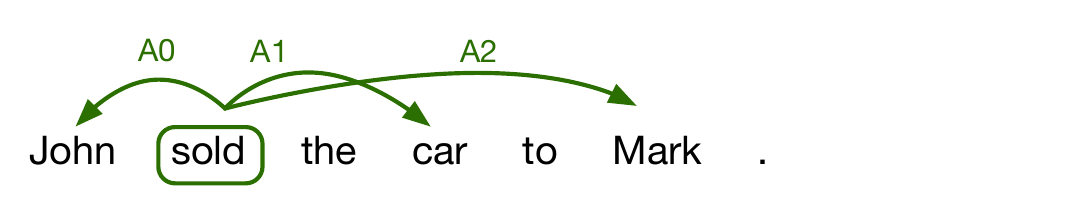}
 \\
BiRNN  & John verkaufte das Auto nach Mark . \\
Sem	&  John verkaufte das Auto an Mark .	\\
\midrule
 & \includegraphics[width=1.2 \columnwidth, height=3.80cm, keepaspectratio]{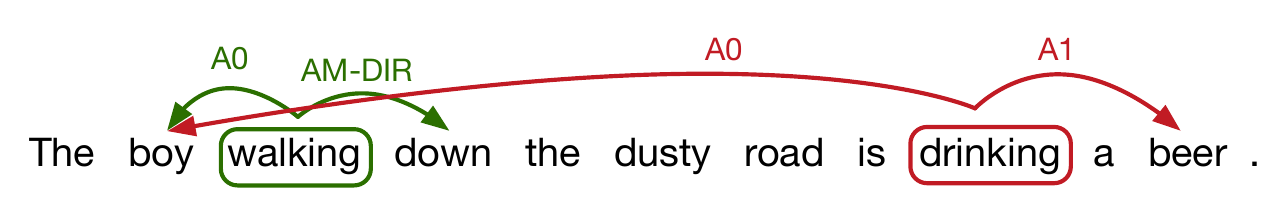} \\
BiRNN  & Der Junge zu Fu{\ss} die staubige Stra{\ss}e ist ein Bier trinken . \\
Sem 	&  Der Junge , der die staubige Stra{\ss}e hinunter geht , trinkt ein Bier .	\\
\midrule
 & \includegraphics[width=1.2 \columnwidth, height=3.80cm, keepaspectratio]{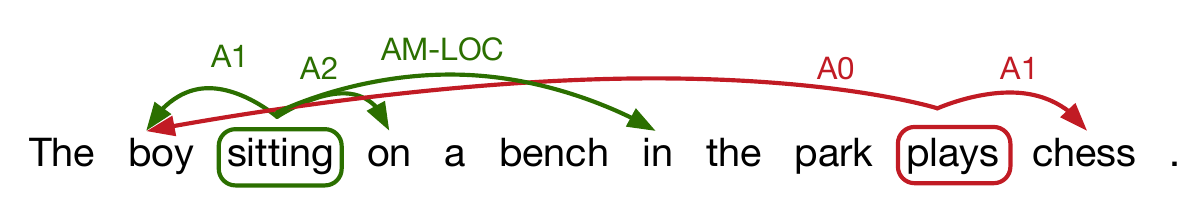} \\
BiRNN  & Der Junge auf einer Bank im Park spielt Schach . \\
Sem 	&  Der Junge sitzt auf einer Bank im Park Schach .	\\
\bottomrule
\end{tabular}
\caption{Qualitative analysis. The first two sentences are translations where the semantic structure helps. For the last sentence both translations are problematic but the BiRNN one is grammatical.}
\label{tab:qualitative}
\end{table*}
\subsection{Ablation and Syntax-Semantics GCNs}

We used the validation set to perform extra experiments, as well as to select hyper parameters (e.g., the number of GCN layers) for the experiments presented above. Table \ref{tab:dev-ende} presents the results. The annotation 1L, 2L and 3L refers to the number of GCN layers used.

First, we tested whether the gain we observed is an effect of an extra layer of non-linearity or an effect of the linguistic structures encoded with GCNs. In order to do so, we used the GCN layer without any structural information.
In this way, only the self-loop edge is used within the GCN node updates.
These results (e.g., BiRNN+SelfLoop) show that the linguistic-agnostic GCNs  perform on par with the baseline, and thus using linguistic structure is genuinely beneficial in translation.

Since syntax and semantic structures seem to be individually beneficial and, though related, capture different linguistic phenomena, it is natural to try combining them.
When syntax and semantic are combined together in the same GCN layer (SemSyn),  we do not observe any improvement with respect to having semantic and syntactic information alone.\footnote{We used distinct  matrices $W$ for syntax and semantics.}  
We argue that the reason for this is that the two linguistic signals do not interact much when encoded into the same GCN layer with a simpler aggregation function.
We thus stacked a semantic GCN on top of a syntactic one and varied the number of layers. 
Though this approach is more successful, we manage to obtain only very moderate improvements over the single-representation models.

\subsection{Qualitative Analysis}
We analyzed the behavior of the BiRNN baseline and the semantic GCN model trained on the full WMT16 training set.
In Table \ref{tab:qualitative} we show three examples where there is a clear difference between translations produced by the two models. Besides the two translations, we show the dependency SRL structure predicted by the labeler and exploited by our GCN model. 

In the first sentence, the only difference is in the choice of the preposition for the argument \texttt{Mark}. Note that the argument is correctly assigned to role \texttt{A2} (`Buyer') by the semantic role labeler. 
The BiRNN model translates \texttt{to} with \texttt{nach}, which in German expresses directionality and would be a correct translation should the argument refer to a location.
In contrast, semantic GCN correctly translates {\tt to} as \texttt{an}.
We hypothesize that the semantic structure, namely the assignment of the argument to {\tt A2} rather than {\tt AM-DIR} (`Directionality'), helps the model to choose the right preposition. 
In the second sentence, the BiRNN's translation is ungrammatical, whereas semantic GCN is able to correctly translate the source sentence. Again, the arguments, correctly identified by semantic role labeler, may have been useful in translating this somewhat tricky sentence.
Finally, in the third case, we can observe that both translations are problematic.
BiRNN and Semantic GCN ignored  verbs {\tt sit} and  {\tt play}, respectively.
However, BiRNN's translation for this sentence is preferable, as it is grammatically correct, even if not fluent or particularly precise.

\section{Conclusions}
In this work we propose injecting information about predicate-argument structures of sentences in NMT models.
We observe that the semantic structures are beneficial for the English--German language pair.
So far we evaluated the model performance in terms of BLEU only. 
It would be interesting in future work to both understand when semantics appears beneficial, and also to see which components of semantic structures play a role.
Experiments on other language pairs are also left for future work.

\section*{Acknowledgments}
We thank Stella Frank and Wilker Aziz for their suggestions and comments.
The project was supported by the European Research Council (ERC StG BroadSem 678254), and the Dutch National Science Foundation (NWO VIDI 639.022.518). 
We thank NVIDIA for donating the GPUs used for this research.

\bibliography{naaclhlt2018}

\begin{thebibliography}{}
\expandafter\ifx\csname natexlab\endcsname\relax\def\natexlab#1{#1}\fi

\bibitem[{Aharoni and Goldberg(2017)}]{aharonigoldberg2017stringtotree}
Roee Aharoni and Yoav Goldberg. 2017.
\newblock \href{https://doi.org/10.18653/v1/P17-2021}{Towards string-to-tree
  neural machine translation}.
\newblock In {\em Proceedings of the 55th Annual Meeting of the Association for
  Computational Linguistics, {ACL}\/}. pages 132--140.
\newblock \url{https://doi.org/10.18653/v1/P17-2021}.

\bibitem[{Andor et~al.(2016)Andor, Alberti, Weiss, Severyn, Presta, Ganchev,
  Petrov, and Collins}]{P16-1231}
Daniel Andor, Chris Alberti, David Weiss, Aliaksei Severyn, Alessandro Presta,
  Kuzman Ganchev, Slav Petrov, and Michael Collins. 2016.
\newblock \href{https://doi.org/10.18653/v1/P16-1231}{Globally normalized
  transition-based neural networks}.
\newblock In {\em Proceedings of the 54th Annual Meeting of the Association for
  Computational Linguistics, {ACL}\/}. pages 2442--2452.
\newblock \url{https://doi.org/10.18653/v1/P16-1231}.

\bibitem[{Aziz et~al.(2011)Aziz, Rios, and Specia}]{DBLP:conf/wmt/AzizRS11}
Wilker Aziz, Miguel Rios, and Lucia Specia. 2011.
\newblock
  \href{http://aclanthology.info/papers/W11-2136/shallow-semantic-trees-for-smt}{Shallow
  semantic trees for {SMT}}.
\newblock In {\em Proceedings of the Sixth Workshop on Statistical Machine
  Translation, WMT@EMNLP\/}. pages 316--322.
\newblock
  \url{http://aclanthology.info/papers/W11-2136/shallow-semantic-trees-for-smt}.

\bibitem[{Bahdanau et~al.(2015)Bahdanau, Cho, and Bengio}]{bahdanau15iclr}
Dzmitry Bahdanau, Kyunghyun Cho, and Yoshua Bengio. 2015.
\newblock \href{http://arxiv.org/abs/1409.0473}{{Neural Machine Translation by
  Jointly Learning to Align and Translate}}.
\newblock In {\em {Proceedings of the International Conference on Learning
  Representations, ICLR}\/}.
\newblock \url{http://arxiv.org/abs/1409.0473}.

\bibitem[{Baker et~al.(2012)Baker, Bloodgood, Dorr, Callison{-}Burch, Filardo,
  Piatko, Levin, and Miller}]{baker-etal:2012:CL}
Kathrin Baker, Michael Bloodgood, Bonnie~J. Dorr, Chris Callison{-}Burch,
  Nathaniel~Wesley Filardo, Christine~D. Piatko, Lori~S. Levin, and Scott
  Miller. 2012.
\newblock \href{https://doi.org/10.1162/COLI_a_00099}{Modality and negation in
  {SIMT} use of modality and negation in semantically-informed syntactic {MT}}.
\newblock {\em Computational Linguistics\/} 38(2):411--438.
\newblock \url{https://doi.org/10.1162/COLI_a_00099}.

\bibitem[{Bar-Hillel(1960)}]{bar1960present}
Yehoshua Bar-Hillel. 1960.
\newblock The present status of automatic translation of languages.
\newblock {\em Advances in Computers\/} 1:91--163.

\bibitem[{Bastings et~al.(2017)Bastings, Titov, Aziz, Marcheggiani, and
  Simaan}]{bastings-EtAl:2017:EMNLP2017}
Jasmijn Bastings, Ivan Titov, Wilker Aziz, Diego Marcheggiani, and Khalil
  Simaan. 2017.
\newblock \href{https://www.aclweb.org/anthology/D17-1209}{Graph convolutional
  encoders for syntax-aware neural machine translation}.
\newblock In {\em Proceedings of the 2017 Conference on Empirical Methods in
  Natural Language Processing, {EMNLP}\/}. pages 1957--1967.
\newblock \url{https://www.aclweb.org/anthology/D17-1209}.

\bibitem[{Bazrafshan and Gildea(2013)}]{DBLP:conf/acl/BazrafshanG13}
Marzieh Bazrafshan and Daniel Gildea. 2013.
\newblock \href{http://aclweb.org/anthology/P/P13/P13-2074.pdf}{Semantic roles
  for string to tree machine translation}.
\newblock In {\em Proceedings of the 51st Annual Meeting of the Association for
  Computational Linguistics, {ACL}\/}. pages 419--423.
\newblock \url{http://aclweb.org/anthology/P/P13/P13-2074.pdf}.

\bibitem[{Cho et~al.(2014)Cho, van Merrienboer, Gulcehre, Bahdanau, Bougares,
  Schwenk, and Bengio}]{cho14emnlp}
Kyunghyun Cho, Bart van Merrienboer, Caglar Gulcehre, Dzmitry Bahdanau, Fethi
  Bougares, Holger Schwenk, and Yoshua Bengio. 2014.
\newblock \href{http://www.aclweb.org/anthology/D14-1179}{{Learning Phrase
  Representations using RNN Encoder--Decoder for Statistical Machine
  Translation}}.
\newblock In {\em {Proceedings of the 2014 Conference on Empirical Methods in
  Natural Language Processing, EMNLP}\/}. pages 1724--1734.
\newblock \url{http://www.aclweb.org/anthology/D14-1179}.

\bibitem[{Elman(1990)}]{elman1990finding}
Jeffrey~L Elman. 1990.
\newblock {Finding structure in time}.
\newblock {\em Cognitive science\/} 14(2):179--211.

\bibitem[{Eriguchi et~al.(2016)Eriguchi, Hashimoto, and
  Tsuruoka}]{eriguchi2016treetoseq}
Akiko Eriguchi, Kazuma Hashimoto, and Yoshimasa Tsuruoka. 2016.
\newblock \href{http://www.aclweb.org/anthology/P16-1078}{Tree-to-sequence
  attentional neural machine translation}.
\newblock In {\em Proceedings of the 54th Annual Meeting of the Association for
  Computational Linguistics, {ACL}\/}. pages 823--833.
\newblock \url{http://www.aclweb.org/anthology/P16-1078}.

\bibitem[{Gehring et~al.(2017)Gehring, Auli, Grangier, and
  Dauphin}]{gehring2016convolutional}
Jonas Gehring, Michael Auli, David Grangier, and Yann Dauphin. 2017.
\newblock \href{https://doi.org/10.18653/v1/P17-1012}{A convolutional encoder
  model for neural machine translation}.
\newblock In {\em Proceedings of the 55th Annual Meeting of the Association for
  Computational Linguistics, {ACL}\/}. pages 123--135.
\newblock \url{https://doi.org/10.18653/v1/P17-1012}.

\bibitem[{Gilmer et~al.(2017)Gilmer, Schoenholz, Riley, Vinyals, and
  Dahl}]{DBLP:conf/icml/GilmerSRVD17}
Justin Gilmer, Samuel~S. Schoenholz, Patrick~F. Riley, Oriol Vinyals, and
  George~E. Dahl. 2017.
\newblock \href{http://proceedings.mlr.press/v70/gilmer17a.html}{Neural message
  passing for quantum chemistry}.
\newblock In {\em Proceedings of the 34th International Conference on Machine
  Learning, {ICML}\/}. pages 1263--1272.
\newblock \url{http://proceedings.mlr.press/v70/gilmer17a.html}.

\bibitem[{Hajic et~al.(2009)Hajic, Ciaramita, Johansson, Kawahara,
  Mart{\'{\i}}, M{\`{a}}rquez, Meyers, Nivre, Pad{\'{o}}, Step{\'{a}}nek,
  Stran{\'{a}}k, Surdeanu, Xue, and
  Zhang}]{DBLP:conf/conll/HajicCJKMMMNPSSSXZ09}
Jan Hajic, Massimiliano Ciaramita, Richard Johansson, Daisuke Kawahara,
  Maria~Ant{\`{o}}nia Mart{\'{\i}}, Llu{\'{\i}}s M{\`{a}}rquez, Adam Meyers,
  Joakim Nivre, Sebastian Pad{\'{o}}, Jan Step{\'{a}}nek, Pavel Stran{\'{a}}k,
  Mihai Surdeanu, Nianwen Xue, and Yi~Zhang. 2009.
\newblock \href{http://aclweb.org/anthology/W/W09/W09-1201.pdf}{The conll-2009
  shared task: Syntactic and semantic dependencies in multiple languages}.
\newblock In {\em Proceedings of the Thirteenth Conference on Computational
  Natural Language Learning: Shared Task, CoNLL\/}. pages 1--18.
\newblock \url{http://aclweb.org/anthology/W/W09/W09-1201.pdf}.

\bibitem[{He et~al.(2016)He, Zhang, Ren, and Sun}]{he2016deep}
Kaiming He, Xiangyu Zhang, Shaoqing Ren, and Jian Sun. 2016.
\newblock \href{https://doi.org/10.1109/CVPR.2016.90}{Deep residual learning
  for image recognition}.
\newblock In {\em Proceedings of the {IEEE} Conference on Computer Vision and
  Pattern Recognition, {CVPR}\/}. pages 770--778.
\newblock \url{https://doi.org/10.1109/CVPR.2016.90}.

\bibitem[{Helcl and Libovick{\'{y}}(2017)}]{NeuralMonkey:2017}
Jind{\v{r}}ich Helcl and Jind{\v{r}}ich Libovick{\'{y}}. 2017.
\newblock \href{https://doi.org/10.1515/pralin-2017-0001}{Neural monkey: An
  open-source tool for sequence learning}.
\newblock {\em The Prague Bulletin of Mathematical Linguistics\/} (107):5--17.
\newblock \url{https://doi.org/10.1515/pralin-2017-0001}.

\bibitem[{Isabelle et~al.(2017)Isabelle, Cherry, and
  Foster}]{DBLP:conf/emnlp/IsabelleCF17}
Pierre Isabelle, Colin Cherry, and George~F. Foster. 2017.
\newblock \href{https://aclanthology.info/papers/D17-1263/d17-1263}{A challenge
  set approach to evaluating machine translation}.
\newblock In {\em Proceedings of the 2017 Conference on Empirical Methods in
  Natural Language Processing, {EMNLP}\/}. pages 2486--2496.
\newblock \url{https://aclanthology.info/papers/D17-1263/d17-1263}.

\bibitem[{Jones et~al.(2012)Jones, Andreas, Bauer, Hermann, and
  Knight}]{C12-1083}
Bevan Jones, Jacob Andreas, Daniel Bauer, Karl~Moritz Hermann, and Kevin
  Knight. 2012.
\newblock \href{http://aclweb.org/anthology/C/C12/C12-1083.pdf}{Semantics-based
  machine translation with hyperedge replacement grammars}.
\newblock In {\em Proceedings of the 24th International Conference on
  Computational Linguistics, {COLING}\/}. pages 1359--1376.
\newblock \url{http://aclweb.org/anthology/C/C12/C12-1083.pdf}.

\bibitem[{Kingma and Ba(2015)}]{kingma2015adam}
Diederik~P. Kingma and Jimmy Ba. 2015.
\newblock \href{http://arxiv.org/abs/1412.6980}{Adam: {A} method for stochastic
  optimization}.
\newblock In {\em Proceedings of the International Conference on Learning
  Representations, ICLR\/}.
\newblock \url{http://arxiv.org/abs/1412.6980}.

\bibitem[{Kipf and Welling(2016)}]{kipf2016semigraphconv}
Thomas~N. Kipf and Max Welling. 2016.
\newblock \href{http://arxiv.org/abs/1609.02907}{Semi-supervised classification
  with graph convolutional networks}.
\newblock In {\em {Proceedings of the International Conference on Learning
  Representations, ICLR}\/}.
\newblock \url{http://arxiv.org/abs/1609.02907}.

\bibitem[{LeCun et~al.(2001)LeCun, Bottou, Bengio, and Haffner}]{lecun-01a}
Yann LeCun, Leon Bottou, Yoshua Bengio, and Patrick Haffner. 2001.
\newblock Gradient-based learning applied to document recognition.
\newblock In {\em Proceedings of Intelligent Signal Processing\/}.

\bibitem[{Linzen et~al.(2016)Linzen, Dupoux, and
  Goldberg}]{linzen-dupoux-goldberg:2016:tacllstmsyntax}
Tal Linzen, Emmanuel Dupoux, and Yoav Goldberg. 2016.
\newblock
  \href{https://www.transacl.org/ojs/index.php/tacl/article/view/972}{Assessing
  the ability of lstms to learn syntax-sensitive dependencies}.
\newblock {\em Transactions of the Association for Computational Linguistics\/}
  4:521--535.
\newblock \url{https://www.transacl.org/ojs/index.php/tacl/article/view/972}.

\bibitem[{Liu and Gildea(2010)}]{DBLP:conf/coling/LiuG10}
Ding Liu and Daniel Gildea. 2010.
\newblock \href{http://aclweb.org/anthology/C10-1081}{Semantic role features
  for machine translation}.
\newblock In {\em Proceedings of the23rd International Conference on
  Computational Linguistics, {COLING}\/}. pages 716--724.
\newblock \url{http://aclweb.org/anthology/C10-1081}.

\bibitem[{Marcheggiani et~al.(2017)Marcheggiani, Frolov, and
  Titov}]{marcheggiani-frolov-titov:2017:srl}
Diego Marcheggiani, Anton Frolov, and Ivan Titov. 2017.
\newblock \href{https://doi.org/10.18653/v1/K17-1041}{A simple and accurate
  syntax-agnostic neural model for dependency-based semantic role labeling}.
\newblock In {\em Proceedings of the 21st Conference on Computational Natural
  Language Learning, {CoNLL}\/}. pages 411--420.
\newblock \url{https://doi.org/10.18653/v1/K17-1041}.

\bibitem[{Marcheggiani and Titov(2017)}]{marcheggiani-titov:2017:srlgcn}
Diego Marcheggiani and Ivan Titov. 2017.
\newblock \href{https://aclanthology.info/papers/D17-1159/d17-1159}{Encoding
  sentences with graph convolutional networks for semantic role labeling}.
\newblock In {\em Proceedings of the 2017 Conference on Empirical Methods in
  Natural Language Processing, {EMNLP}\/}. pages 1506--1515.
\newblock \url{https://aclanthology.info/papers/D17-1159/d17-1159}.

\bibitem[{Nadejde et~al.(2017)Nadejde, Reddy, Sennrich, Dwojak,
  Junczys{-}Dowmunt, Koehn, and Birch}]{DBLP:conf/wmt/NadejdeRSDJKB17}
Maria Nadejde, Siva Reddy, Rico Sennrich, Tomasz Dwojak, Marcin
  Junczys{-}Dowmunt, Philipp Koehn, and Alexandra Birch. 2017.
\newblock \href{http://aclanthology.info/papers/W17-4707/w17-4707}{Predicting
  target language {CCG} supertags improves neural machine translation}.
\newblock In {\em Proceedings of the Second Conference on Machine Translation,
  {WMT}\/}. pages 68--79.
\newblock \url{http://aclanthology.info/papers/W17-4707/w17-4707}.

\bibitem[{Palmer et~al.(2005)Palmer, Kingsbury, and
  Gildea}]{DBLP:journals/coling/PalmerKG05}
Martha Palmer, Paul Kingsbury, and Daniel Gildea. 2005.
\newblock \href{https://doi.org/10.1162/0891201053630264}{The proposition bank:
  An annotated corpus of semantic roles}.
\newblock {\em Computational Linguistics\/} 31(1):71--106.
\newblock \url{https://doi.org/10.1162/0891201053630264}.

\bibitem[{Papineni et~al.(2002)Papineni, Roukos, Ward, and
  Zhu}]{papineni2002bleu}
Kishore Papineni, Salim Roukos, Todd Ward, and Wei{-}Jing Zhu. 2002.
\newblock \href{http://www.aclweb.org/anthology/P02-1040.pdf}{Bleu: a method
  for automatic evaluation of machine translation}.
\newblock In {\em Proceedings of the 40th Annual Meeting of the Association for
  Computational Linguistics, {ACL}\/}. pages 311--318.
\newblock \url{http://www.aclweb.org/anthology/P02-1040.pdf}.

\bibitem[{Scarselli et~al.(2009)Scarselli, Gori, Tsoi, Hagenbuchner, and
  Monfardini}]{scarselli2009graph}
Franco Scarselli, Marco Gori, Ah~Chung Tsoi, Markus Hagenbuchner, and Gabriele
  Monfardini. 2009.
\newblock \href{https://doi.org/10.1109/TNN.2008.2005605}{The graph neural
  network model}.
\newblock {\em {IEEE} Trans. Neural Networks\/} 20(1):61--80.
\newblock \url{https://doi.org/10.1109/TNN.2008.2005605}.

\bibitem[{Sennrich and Haddow(2016)}]{sennrich2016linguistic}
Rico Sennrich and Barry Haddow. 2016.
\newblock \href{http://www.aclweb.org/anthology/W16-2209}{{Linguistic Input
  Features Improve Neural Machine Translation}}.
\newblock In {\em {Proceedings of the First Conference on Machine Translation,
  WMT}\/}. pages 83--91.
\newblock \url{http://www.aclweb.org/anthology/W16-2209}.

\bibitem[{Sennrich et~al.(2016)Sennrich, Haddow, and
  Birch}]{sennrich2016subword}
Rico Sennrich, Barry Haddow, and Alexandra Birch. 2016.
\newblock \href{http://www.aclweb.org/anthology/P16-1162}{Neural machine
  translation of rare words with subword units}.
\newblock In {\em Proceedings of the 54th Annual Meeting of the Association for
  Computational Linguistics, {ACL}\/}. pages 1715--1725.
\newblock \url{http://www.aclweb.org/anthology/P16-1162}.

\bibitem[{Surdeanu et~al.(2008)Surdeanu, Johansson, Meyers, M{\`{a}}rquez, and
  Nivre}]{DBLP:conf/conll/SurdeanuJMMN08}
Mihai Surdeanu, Richard Johansson, Adam Meyers, Llu{\'{\i}}s M{\`{a}}rquez, and
  Joakim Nivre. 2008.
\newblock \href{http://aclweb.org/anthology/W/W08/W08-2121.pdf}{The {CoNLL}
  2008 shared task on joint parsing of syntactic and semantic dependencies}.
\newblock In {\em Proceedings of the Twelfth Conference on Computational
  Natural Language Learning, CoNLL\/}. pages 159--177.
\newblock \url{http://aclweb.org/anthology/W/W08/W08-2121.pdf}.

\bibitem[{Vaswani et~al.(2017)Vaswani, Shazeer, Parmar, Uszkoreit, Jones,
  Gomez, Kaiser, and Polosukhin}]{NIPS2017_7181}
Ashish Vaswani, Noam Shazeer, Niki Parmar, Jakob Uszkoreit, Llion Jones,
  Aidan~N. Gomez, Lukasz Kaiser, and Illia Polosukhin. 2017.
\newblock
  \href{http://papers.nips.cc/paper/7181-attention-is-all-you-need}{Attention
  is all you need}.
\newblock In {\em Advances in Neural Information Processing Systems 30: Annual
  Conference on Neural Information Processing Systems, NIPS\/}. pages
  6000--6010.
\newblock \url{http://papers.nips.cc/paper/7181-attention-is-all-you-need}.

\bibitem[{Weaver(1955)}]{weaver1955translation}
Warren Weaver. 1955.
\newblock Translation.
\newblock {\em Machine translation of languages\/} 14:15--23.

\bibitem[{Wu and Fung(2009)}]{DBLP:conf/naacl/WuF09}
Dekai Wu and Pascale Fung. 2009.
\newblock \href{http://www.aclweb.org/anthology/N09-2004}{Semantic roles for
  {SMT:} {A} hybrid two-pass model}.
\newblock In {\em Proceedings of the Human Language Technologies: Conference of
  the North American Chapter of the Association of Computational Linguistics,
  {NAACL}\/}. pages 13--16.
\newblock \url{http://www.aclweb.org/anthology/N09-2004}.

\end{thebibliography}
\bibliographystyle{acl_natbib}

\appendix
\newpage

\newpage

\section{Hyperparameters}
\label{sec:hyperparameters}
For experiments on the News Commentary data we used 8000 BPE merges, whereas we used 16000 BPE merges for En--De experiments on the full dataset.
For all the experiments, we used bidirectional GRUs and we set the embedding size to 256, we used word dropout with retain probability of 0.8 and edge dropout with the same probability, we used L2 regularization on all the parameters with value of $10^{-8}$, translations are obtained using a greedy decoder.
We placed residual connections \cite{he2016deep} before every GCN layer.
For the experiments on News Commentary data, we set GRU (for both encoder and decoder) and CNN hidden states to 512, we use Adam \cite{kingma2015adam} as optimizer with an initial learning rate of 0.0002, and we trained the models for 50 epochs.
For large scale experiments on En--De, we set the GRU hidden states to 800, and instead of greedy decoding we employed beam search (beam 12). 
We trained the model for 20 epochs with the same hyperparameters.

\section{Datasets Statistics}
\label{sec:data-stats}
\begin{table}[bh]
\centering
\begin{tabular}{@{}lrrr@{}} 
\toprule
          & Train    & Val. & Test \\ 
\midrule
English--German & 226822 & 2169 & 2999 \\
English--German (full) & 4500966  & 2169 & 2999\\

\bottomrule
\end{tabular}
\caption{The number of sentences in our datasets.}
\label{tab:data}
\end{table}

\begin{table}[th]
\centering
\begin{tabular}{@{}lrr@{}} 
\toprule
           & Source    & Target \\ 
\midrule
English--German & 37824 & 8099 (BPE) \\
English--German (full)      & 50000 & 16000 (BPE) \\
\bottomrule
\end{tabular}
\caption{Vocabulary sizes.}
\label{tab:voc}
\end{table}

\end{document}